\documentclass{llncs}
\usepackage{graphicx}
\usepackage{multirow}
\usepackage{microtype}
\usepackage{booktabs}
\usepackage[hidelinks]{hyperref}
\usepackage[
	draft, 
	textsize=footnotesize, 
    color=green!20, 
    linecolor=black
]{todonotes}

\usepackage{verbatim}

\usepackage{subfig}

\emergencystretch =\maxdimen
\begin{document}

\title{Automatic Detection of Knee Joints and Quantification of Knee Osteoarthritis Severity using Convolutional Neural Networks}

\author{Joseph Antony\inst{1}, Kevin McGuinness\inst{1}, Kieran Moran\inst{1,2} \and Noel E O'Connor\inst{1}
}

\institute{Insight Centre for Data Analytics, Dublin City University, Dublin, Ireland. \inst{1} \\ School of Health and Human Performance, Dublin City University, Dublin, Ireland. \inst{2} \\
\email{joseph.antony@insight-centre.org}}

\maketitle

% ---------------------------------- %

\begin{abstract}
This paper introduces a new approach to automatically quantify the severity of knee OA using X-ray images. Automatically quantifying knee OA severity involves two steps: first, automatically localizing the knee joints; next, classifying the localized knee joint images. We introduce a new approach to automatically detect the knee joints using a fully convolutional neural network (FCN). We train convolutional neural networks (CNN) from scratch to automatically quantify the knee OA severity optimizing a weighted ratio of two loss functions: categorical cross-entropy and mean-squared loss. This joint training further improves the overall quantification of knee OA severity, with the added benefit of naturally producing simultaneous multi-class classification and regression outputs. Two public datasets are used to evaluate our approach, the Osteoarthritis Initiative (OAI) and the Multicenter Osteoarthritis Study (MOST), with extremely promising results that outperform existing approaches. 


\end{abstract}

% ---------------------------------- %

\section{Introduction}

%%% Motivation %%%
Knee Osteoarthritis (OA) is a debilitating joint disorder that mainly degrades the knee articular cartilage. Clinically, the major pathological features for knee OA include joint space narrowing, osteophytes formation, and sclerosis. 
%The causes for knee OA include mechanical abnormalities such as degradation of articular cartilage, menisci, ligaments, synovial tissue and sub-chondral bone. 
Knee OA has a high-incidence among the elderly, obese, and those with a sedentary lifestyle.
%Recently, knee OA is highly incident among elder populace, and the populace not following an active life style and obese.
In its severe stages, it causes excruciating pain and often leads to total joint arthoplasty. Early diagnosis is crucial for clinical treatments and pathology \cite{oka2008fully,shamir2009early}. Despite the introduction of several imaging modalities such as MRI, Optical Coherence Tomography and ultrasound for augmented OA diagnosis, radiography (X-ray) has been traditionally preferred, and remains the main accessible tool and ``gold standard" for preliminary knee OA diagnosis \cite{oka2008fully,shamir2009knee,shamir2008wndchrm}.

%%% Proposed Approach %%%
Previous work has approached automatically assessing knee OA severity \cite{shamir2009early,shamir2008wndchrm,yoo2016simple} as an image classification problem. In this work, we train CNNs from scratch to automatically quantify knee OA severity using X-ray images. This involves two main steps: 1) automatically detecting and extracting the region of interest (ROI) and localizing the knee joints, 2) classifying the localized knee joints.

We introduce a fully-convolutional neural network (FCN) based method to automatically localize the knee joints. A FCN is an end-to-end network trained to make pixel-wise predictions \cite{long2015fully}. Our FCN based method is highly accurate for localizing knee joints and the FCN can easily fit into an end-to-end network trained to quantify knee OA severity. 

To automatically classify the localized knee joints we propose two methods: 1) training a CNN from scratch for multi-class classification of knee OA images, and 2) training a CNN to optimize a weighted ratio of two loss functions: categorical cross-entropy for multi-class classification and mean-squared error for regression. We compare the results from these methods to WND-CHARM \cite{shamir2009knee,shamir2008wndchrm} and our previous study \cite{antony2016quantifying}. We also compare the classification results to both manual and automatic localization of knee joints.

%%% Contributions %%%
We propose a novel pipeline to automatically quantify knee OA severity including a FCN for localizing knee joints and a CNN jointly trained for classification and regression of knee joints. The main contributions of this work include the fully-convolutional network (FCN) based method to automatically localize the knee joints, training a network (CNN) from scratch that optimizes  a weighted ratio of both categorical cross-entropy for multi-class classification and mean-squared error for regression of knee joints. This multi-objective convolutional learning improves the overall quantification with an added benefit of providing simultaneous multi-class classification and regression outputs. 

% ---------------------------------- %

 %%% Related Work %%%
 \section{Related Work}
%Previous work \cite{orlov2008wnd,shamir2009early,shamir2009knee,shamir2008wndchrm} has approached automatically assessing knee OA from radiographs as an image classification problem. 
Assessing knee OA severity through classification can be achieved by detecting the variations in joint space width and osteophytes formation in the knee joints \cite{oka2008fully,shamir2009early,shamir2009knee}. In a recent approach, Yoo et. al. used artificial neural networks (ANN) and KNHANES V-1 data, and developed a scoring system to predict radiographic and symptomatic knee OA~\cite{yoo2016simple} risks. Shamir et. al. used WND-CHARM: a multipurpose bio-medical image classifier \cite{orlov2008wnd} to classify knee OA radiographs \cite{shamir2013wnd,shamir2008wndchrm} and for early detection of knee OA using computer aided analysis \cite{shamir2009early}. WND-CHARM uses hand-crafted features extracted from raw images and image transforms \cite{orlov2008wnd,shamir2013wnd}. 

Recently, convolutional neural networks (CNNs) have outperformed many methods based on hand-crafted features and they are highly successful in many computer vision tasks such as image recognition, automatic detection and segmentation, content based image retrieval, and video classification. CNNs learn effective feature representations particularly well-suited for fine-grained classification \cite{yang2013feature} like classification of knee OA images. In our previous study \cite{antony2016quantifying}, we showed that the off-the-shelf CNNs such as the VGG 16-Layers network \cite{simonyan2014very}, the VGG-M-128 network \cite{chatfield2014return}, and the BVLC reference CaffeNet \cite{jia2014caffe,karayev2013recognizing} trained on ImageNet LSVRC dataset \cite{russakovsky2015imagenet} can be fine-tuned for classifying knee OA images through transfer learning. We also argued that it is appropriate to assess  knee OA severity using a continuous metric like mean-squared error instead of binary or multi-class classification accuracy, and showed that predicting the continuous grades through regression reduces the mean-squared error and in turn improves the overall quantification. 

Previously, Shamir et. al. \cite{shamir2009early} proposed template matching to automatically detect and extract the knee joints. This method is slow for large datasets such as OAI, and the accuracy and precision of detecting knee joints is low. In our previous study, we introduced an SVM-based method for automatically detecting the center of knee joints \cite{antony2016quantifying} and extract a fixed region with reference to the detected center as the ROI. This method is also not highly accurate and there is a compromise in the aspect ratio of the extracted knee joints that affects the overall quantification.

% ---------------------------------- %

\section{Data}
The data used for the experiments and analysis in this study are bilateral PA fixed flexion knee X-ray images. The datasets are from the Osteoarthritis Initiative (OAI) and Multicenter Osteoarthritis Study (MOST) in the University of California, San Francisco, and are standard datasets used in knee osteoarthritis studies.

\subsection{Kellgren and Lawrence Grades}
\begin{figure}[ht]
  \centering
  \includegraphics[scale=0.25]{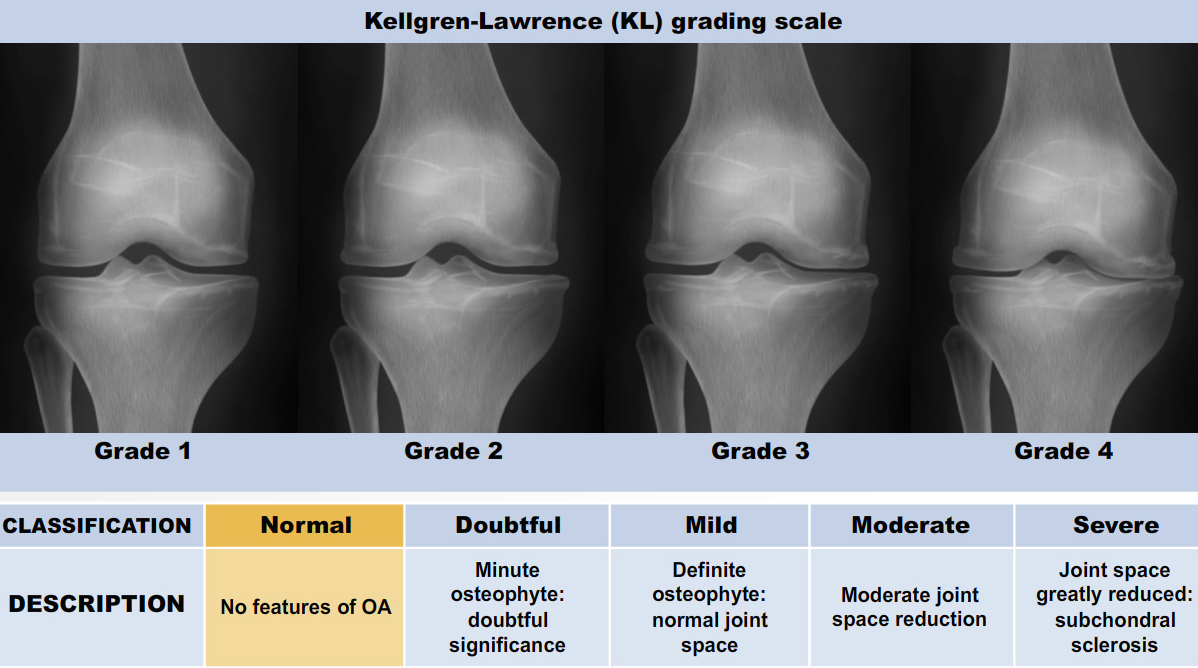}
  \caption{The KL grading system to assess the severity of knee OA.
  	%\footnotesize{Source: \url{http://www.adamondemand.com/clinical-management-of-osteoarthritis/}}
  }
  \label{fig:KL}
\end{figure}

This study uses Kellgren and Lawrence (KL) grades as the ground truth to classify the knee OA X-ray images. The KL grading system is still considered the gold standard for initial assessment of knee osteoarthritis severity in radiographs \cite{oka2008fully,orlov2008wnd,park2013practical,shamir2009knee}. It uses five grades to indicate radiographic knee OA severity. `Grade 0' represents normal, `Grade 1' doubtful, `Grade 2' minimal, `Grade 3' moderate, and `Grade 4' represents severe. Figure  \ref{fig:KL} shows the KL grading system.

\subsection{OAI and MOST Data Sets}
The baseline cohort of the OAI dataset contains MRI and X-ray images of 4,476 participants. From this entire cohort, we selected 4,446 X-ray images based on the availability of KL grades for both knees as per the assessments by Boston University X-ray reading center (BU). In total there are 8,892 knee images and the distribution as per the KL grades is as follows: Grade 0 - 3433, Grade 1 - 1589, Grade 2 - 2353, Grade 3 - 1222, and Grade 4 - 295.

%\subsection{MOST Dataset}
The MOST dataset includes lateral knee radiograph assessments of 3,026 participants. From this, 2,920 radiographs are selected based on the availability of KL grades for both knees as per baseline to 84-month Longitudinal Knee Radiograph Assessments. In this dataset there are 5,840 knee images and the distribution as per KL grades is as follows: Grade 0 - 2498, Grade 1 - 1018, Grade 2 - 923, Grade 3 - 971, and Grade 4 - 430.

% ---------------------------------- %

\section{Methods}
\begin{figure}[ht]
  \centering
  \includegraphics[scale=0.45]{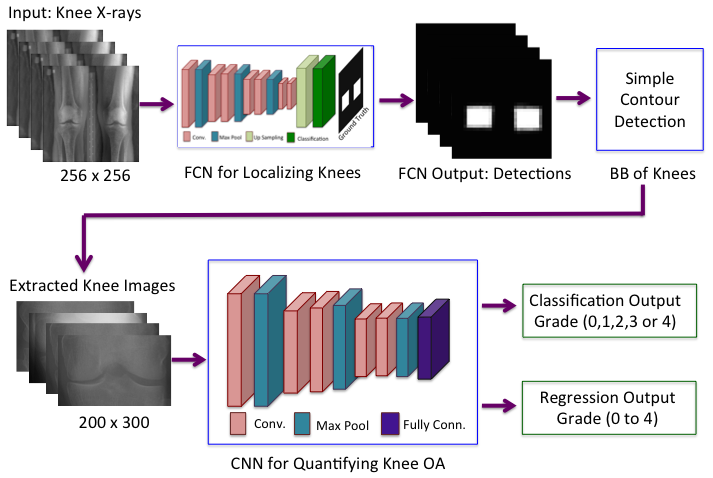}
  \caption{The pipeline used for quantifying knee OA severity.}
  \label{fig:Pipeline}
\end{figure}

This section introduces the methodology used for quantifying radiographic knee OA severity. This involves two steps: automatically detecting knee joints using a fully convolutional network (FCN), and simultaneous classification and regression of localized knee images using a convolutional neural network (CNN). Figure \ref{fig:Pipeline} shows the complete pipeline used for quantifying knee OA severity.  

\subsection{Automatically Localizing Knee Joints using a FCN}
\begin{figure}[ht]
  \centering
  \includegraphics[scale = 0.2]{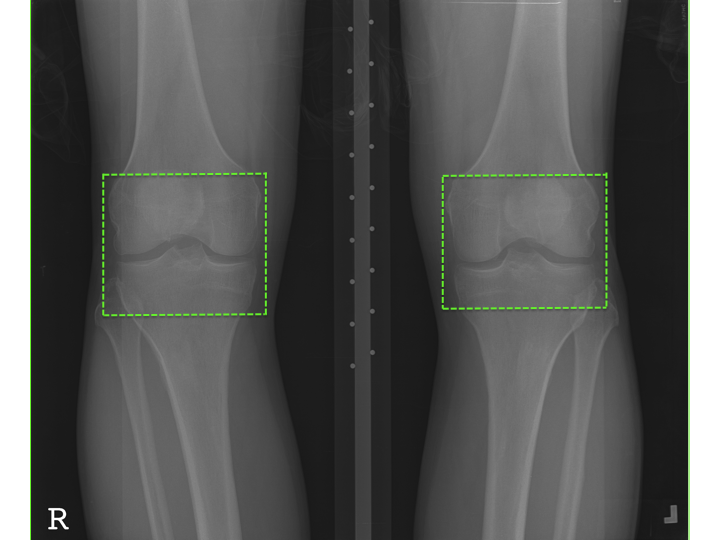}
  \caption{A knee OA X-ray image with the region of interest: the knee joints.}
  \label{fig:ROI}
\end{figure}

Assessment of knee OA severity can be achieved by detecting the variations in joint space width and osteophytes formation in the knee joint~\cite{oka2008fully}. Thus, localizing the knee joints from the X-ray images is an essential pre-processing step before quantifying knee OA severity, and for larger datasets automatic methods are preferable. Figure \ref{fig:ROI} shows a knee OA radiograph and the knee joints: the region of interest (ROI) for detection. The previous methods for automatically localizing knee joints such as template matching \cite{shamir2009early} and our own SVM-based method \cite{antony2016quantifying} are not very accurate. In this study, we propose a fully convolutional neural network (FCN) based approach to further improve the accuracy and precision of detecting knee joints. %and also to develop a method that is suitable to integrate into an end-to-end deep learning system for quantifying knee OA severity.\todo{Clarify what you mean by end to end. It often means that you can train the whole system just given the answer. In this case, you can't (you need detection ground truth, and cant backprop grads for the full end-to-end system)}

\subsubsection{FCN Architecture:}
\begin{figure}[ht]
  \centering
  \includegraphics[scale=0.5]{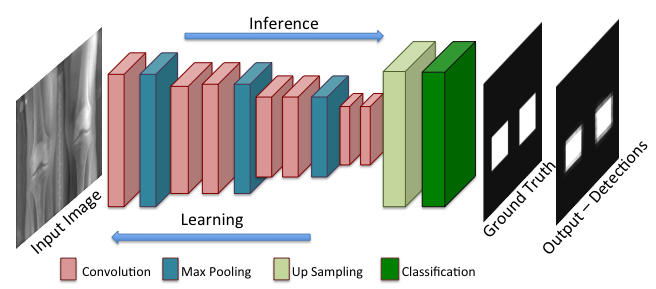}
  \caption{The Fully Convolutional Network for automatically detecting knee joints.}
  \label{fig:AD_FCN}
\end{figure}

Inspired by the success of a fully convolutional neural network (FCN) for semantic segmentation on general images \cite{long2015fully}, we trained a FCN to automatically detect the region of interest (ROI): the knee joints from the knee OA radiographs. Our proposed FCN is based on a lightweight architecture and the network parameters are trained from scratch. Figure \ref{fig:AD_FCN} shows the architecture. After experimentation, we found this architecture to be the best for knee joint detection. The network consists of 4 stages of convolutions with a max-pooling layer after each convolutional stage, and the final stage of convolutions is followed by an up-sampling and a fully-convolutional layer. The first and second stages of convolution use 32 filters, the third stage uses 64 filters, and the fourth stage uses 96 filters. The network uses a uniform [$3\times3$] convolution and [$2\times2$] max pooling. Each convolution layer is followed by a batch normalization and a rectified linear unit activation layer (ReLU). After the final convolution layer, an [$8\times8$] up-sampling is performed as the network uses 3 stages of [$2\times2$] max pooling. The up-sampling is essential for an end-to-end learning by back propagation from the pixel-wise loss and to obtain pixel-dense outputs \cite{long2015fully}. The final layer is a fully convolutional layer with a kernel size of [$1\times1$] and uses a sigmoid activation for pixel-based classification. The input to the network is of size [$256\times256$] and the output is of same size.

\subsubsection{FCN Training:} 
We trained the network from scratch with  training samples of knee OA radiographs from the OAI and MOST datasets. The ground truth for training the network are binary images with masks specifying the ROI: the knee joints. Figure \ref{fig:AD_FCN} shows an instance of the binary masks: the ground truth. We generated the binary masks from manual annotations of knee OA radiographs using a fast annotation tool that we developed. The network was trained to minimize the total binary cross entropy between the predicted pixels and the ground truth. We used the adaptive moment estimation (Adam) optimizer \cite{kingma2014adam}, with default parameters, which we found to give faster convergence than standard SGD.

\subsubsection{Extracting Knee Joints:}
We deduce the bounding boxes of the knee joints using simple contour detection from the output predictions of FCN. We extract the knee joints from knee OA radiographs using the bounding boxes. We upscale the bounding boxes from the output of the FCN that is of size [$256\times256$] to the original size of each knee OA radiograph before we extract the knee joints so that the aspect ratio of the knee joints is preserved.

\subsection{Quantifying knee OA severity using CNNs}
We investigate the use of CNNs trained from scratch using knee OA data and jointly train networks to minimize the classification and regression losses to further improve the assessment of knee OA severity.

\subsubsection{Training CNN for Classification:}
\begin{figure}[ht]
  \centering
  \includegraphics[scale=0.5]{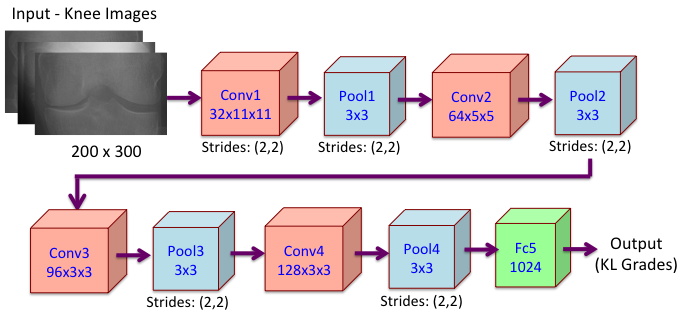}
  \caption{The network architecture for classifying knee joint images.}
  \label{fig:ClsfArch}
\vspace{-5mm}
\end{figure}

The network contains mainly five layers of learned weights: four convolutional layers and one fully connected layer. Figure \ref{fig:ClsfArch} shows the network architecture. As the training data is relatively scarce, we considered a lightweight architecture with minimal layers and the network has 5.4 million free parameters in total. After experimenting with the number of convolutional layers and other parameters, we find this architecture to be the best for classifying knee images. Each convolutional layer in the network is followed by batch normalization and a rectified linear unit activation layer (ReLU). After each convolutional stage there is a max pooling layer. The final pooling layer is followed by a fully connected layer and a softmax dense layer. To avoid over-fitting, we include a drop out layer with a drop out ratio of 0.2 after the last convolutional (conv4) layer and a drop out layer with a drop out ratio of 0.5 after the fully connected layer (fc5). We also apply an L2-norm weight regularization penalty of 0.01 in the last two convolutional layers (conv3 and conv4) and the fully connected layer (fc5). Applying a regularization penalty to other layers increases the training time whilst not introducing significant variation in the learning curves. The network was trained to minimize categorical cross-entropy loss using the Adam optimizer \cite{kingma2014adam}. The inputs to the network are knee images of size [200$\times$300]. We chose this size to approximately preserve the aspect ratio  based on the mean aspect ratio (1.6) of all the extracted knee joints.

\subsubsection{Jointly training CNN for Classification and Regression:}
\begin{figure}[ht]
  \centering
  \includegraphics[scale=0.5]{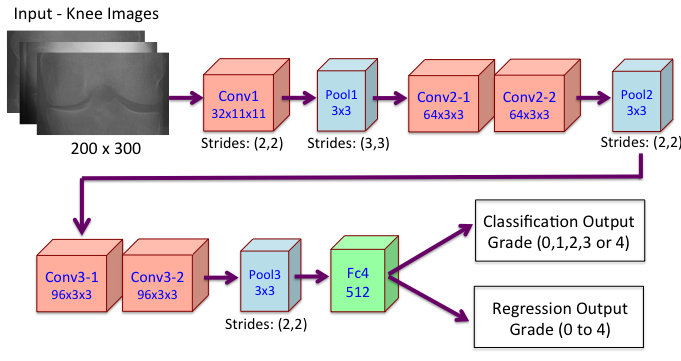}
  \caption{The network architecture for simultaneous classification and regression.}
  \label{fig:JointArch}
\vspace{-5mm}
\end{figure}

In general, assessing knee OA severity is based on the multi-class classification of knee images and assigning KL grade to each distinct category \cite{oka2008fully,orlov2008wnd,shamir2009early,shamir2008wndchrm}. As the disease is progressive in nature, we argued in our previous paper \cite{antony2016quantifying} that assigning a continuous grade (0--4) to knee images through regression is a better approach for quantifying knee OA severity. However, with this approach there is no ground truth of KL grades in a continuous scale  to train a network directly for regression output. Therefore, we train networks using multi-objective convolutional learning~ \cite{liu2015multi} to optimize a weighted-ratio of two loss functions: categorical cross-entropy and mean-squared error. Mean squared error gives the network information about ordering of grades, and cross entropy gives information about the quantization of grades. Intuitively, optimizing a network with two loss functions provides a stronger error signal and it is a step to improve the overall quantification, considering both classification and regression results. After experimenting, we obtained the final architecture shown in Figure \ref{fig:JointArch}. This network has six layers of learned weights: 5 convolutional layers and a fully connected layer, and approximately 4 million free parameters in total. Each convolutional layer is followed by batch normalization and a rectified linear activation (ReLU) layer. To avoid over-fitting this model, we include drop out ($p=0.5$) in the fully connected layer (fc5) and L2 weight regularization in the fully connected layer (fc5) and the last stage of convolution layers (Conv3-1 and Conv3-2). We trained the model using stochastic gradient descent with \textit{Nesterov} momentum and a learning rate scheduler. The initial learning rate was set to 0.001, and reduced by a factor of 10 if there is no drop in the validation loss for 4 consecutive epochs. 

% ---------------------------------- %

\section{Experiments and Results}
%In this section, we discuss about the experiments and implementations for localizing the knee joints using fully convolutional neural network (FCN) and quantifying knee OA severity through classification and regression using convolutional nets (CNN), and we analyze the results. 
%\todo[inline]{You can probably drop these preamble texts at the start of each section, since they don't add much more than the title.}

\subsection{Localizing the Knee Joints using a FCN}
We trained FCNs to automatically localize and extract the knee joints from knee OA X-ray images. We use the well-known Jaccard index to evaluate the detection result. The datasets are split into a training/validation set (70\%) and test set (30\%). The training and test samples from OAI dataset are 3,146 images and 1,300 images. The training and test samples from MOST dataset are 2,020 images and 900 images. First, we trained the network with training samples from OAI dataset and tested it with OAI and MOST datasets separately. Next, we increased our training samples by including the MOST training set and the test set is a combination of both OAI and MOST test sets. Before settling on the final architecture, we experimented by varying the number of convolution stages, the number of filters and kernel sizes in each convolution layer. The final network (shown in Figure \ref{fig:AD_FCN}) was trained with the samples from both OAI and MOST datasets.

\begin{comment}
\begin{figure}[ht]
  \centering
  \includegraphics[scale=0.75]{FCN_LearningCurves.png}
  \caption{Learning Curves of FCN trained for localizing knee joints.}
  \label{fig:FCN_LC}
\end{figure}
\end{comment}

\subsubsection{Evaluation:} The automatic detection is evaluated using the well-known Jaccard index i.e. the intersection over Union (IoU) of the automatic detection and the manual annotation of each knee joint. For this evaluation, we manually annotated all the knee joints in both the OAI and MOST datasets using a fast annotation tool that we developed. Table \ref{Tab:AD} shows the number (percentage) of knee joint correctly detected based on the Jaccard index (J) values greater than 0.25, 0.5 and 0.75 along with the mean and the standard deviation of J. Table \ref{Tab:AD} also shows detection rates on the OAI and MOST test sets separately.

\begin{table}
\vspace{-5mm}
\caption{Comparison of automatic detection based on the Jaccard Index (J)}
\label{Tab:AD}
\centering
\begin{tabular}{c c c c c c}
\toprule
Test Data & J\ensuremath{\ge}0.25 & J\ensuremath{\ge}0.5 & J\ensuremath{\ge}0.75 & Mean  & Std.Dev \tabularnewline
\midrule
\midrule
OAI & \textbf{100\%} & \textbf{99.9\%}& 89.2\% & 0.83 & 0.06\tabularnewline
MOST & 99.5\% & 98.4\% & 85.0\% & 0.81 & 0.09\tabularnewline
Combined OAI-MOST  & 99.9\% & \textbf{99.9\%} & \textbf{91.4\%} & 0.83 & 0.06\tabularnewline
\bottomrule
\end{tabular}
\vspace{-8mm}
\end{table}

\subsubsection{Results:} Considering the anatomical variations of the knee joints and the imaging protocol variations, the automatic detection with a FCN is highly accurate with 99.9\% (4,396 out of 4,400) of the knee joints for J$\geq$0.5 and 91.4\% (4,020 out of 4,400) of the knee joints for J$\geq$0.75 being correctly detected. %\todo{Include the total number of mis-classifications on the full test set} There is a decrease in detection accuracy (85\%) when the network is trained with OAI data and tested with MOST data. This is due to the variations in the X-ray imaging protocols that can be visually observed \todo{Is the figures for this in the table? If not, include them in the text. Also, what do these variations look like? Can you show an example? Could they be simulated?}. To overcome these variations, we combined the OAI-MOST data set for training the network. 
Section 5.3 gives further evidence that the FCN based detection is highly accurate by showing that the quantification results obtained with the automatically extracted knee joints gives results on par with manually segmented knee joints.

\subsection{Classification of Knee OA Images using a CNN}
We use the same train-test split for localization and quantification to maintain uniformity in the pipeline and to enable valid comparisons of the results obtained across the various approaches.  %before training networks with automatically extracted knee joint images. 
We include the right-left flip of each knee joint image to increase the training samples and this doubles the total number of training samples available. As an initial approach, we trained networks to classify manually annotated knee joint images. After experimenting, we obtained the final architecture shown in Figure \ref{fig:ClsfArch}. %Figure \ref{fig:Clsf_LC} shows the learning curves obtained while training our network. 

\begin{comment}
\end{}
\begin{figure}[ht]
  \centering
  \includegraphics[scale=0.75]{ClsfLossAcc.png}
  \caption{Learning Curves for classifying knee OA images.}
  \label{fig:Clsf_LC}
  \todo[inline]{You have loss and accuracy on the same axis here, but they do not have the same units. Better to put loss on the left side and accuracy on the right.}
\end{figure}
\end{comment}

\subsubsection{Results:} we compare the classification results from our network to WND-CHARM, the multipurpose medical image classifier \cite{orlov2008wnd,shamir2008wndchrm,shamir2013wnd} that gave the previous best results for automatically quantifying knee OA severity. Table \ref{Tab:CompClsf} shows the multi-class classification accuracy and mean-squared error of our network and WND-CHARM. The results show that our network trained from scratch for classifying knee OA images clearly outperforms WND-CHARM. Also these results show an improvement over our earlier reported methods \cite{antony2016quantifying} that used off-the-shelf networks such as VGG nets and the BVLC Reference CaffeNet for classifying knee OA X-ray images through transfer learning. These improvements are due to the lightweight architecture of our network trained from scratch with less (5.4 million) free parameters in comparison to 62 million free parameters of BVLC CaffeNet for the given small amount of training data. The off-the-shelf networks were trained using a large dataset like ImageNet containing  millions of images, whereas our dataset contains much fewer ($\sim10,000$) training samples. We show further improvements in the results for quantifying knee OA severity in the next section. 

\begin{table}[ht]
\vspace{-5mm}
\caption{Classification results of our network and WND-CHARM.}
\label{Tab:CompClsf}
\centering
\begin{tabular}{l c c c}
\toprule
Method & Test Data & Accuracy & Mean-Squared Error\\
\midrule
\midrule
 Wndchrm & OAI  & 29.3\% & 2.496 \\
 Wndchrm & MOST & 34.8\% & 2.112 \\
 Fine-Tuned BVLC CaffeNet & OAI & 57.6 \% & 0.836 \\
 \textbf{Our CNN trained from Scratch} & OAI {\&} MOST & \textbf{60.3\%} & 0.898 \\
\bottomrule
\end{tabular}
\vspace{-8mm}
\end{table}
 
\subsection{Jointly trained CNN for Classification and Regression}
\begin{figure}[!ht]
  \centering
  \subfloat[Accuracy Curves.]{\includegraphics[width=0.5\textwidth]{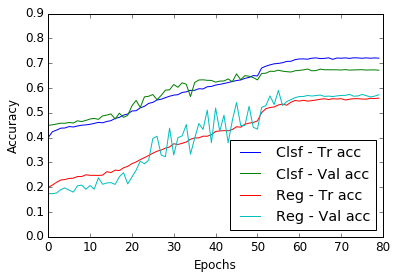}\label{fig:JointAcc}}
  \hfill
  \subfloat[Loss Curves.]{\includegraphics[width=0.5\textwidth]{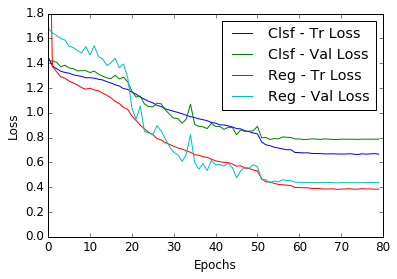}\label{fig:JointLoss}}
  \caption{(a) Training (Tr) and validation (Val) accuracy (acc), (b) Training and validation loss for joint classification (Clsf) and regression (Reg) training.}
  \label{fig:JointLossAcc}
\vspace{-5mm}
\end{figure}

The KL grades used to assess knee OA is a discrete scale, but  knee OA is progressive in nature.  We trained networks to predict the outcomes in a continuous scale (0--4) through regression. Even though we obtained low mean-squared error values for regression, the classification accuracy reduces when the continuous grades are rounded. Next, to obtain a better learning representation we trained networks that learn using a weighted ratio of two loss functions: categorical cross entropy for classification and mean-squared error for regression. We experimented with values from 0.2 to 0.6 for the weight of regression loss and we fixed the weight at 0.5 as this gave the optimal results. Figure \ref{fig:JointArch} shows our network jointly trained for classification and regression of knee images. Figure \ref{fig:JointLossAcc} shows the learning curves of the network trained for joint classification and regression. The learning curves show a decrease in training and validation losses, and also an increase in training and validation accuracies over the training. 

\begin{table}[ht]
\caption{Classification of knee joints after manual and automatic localization.}
\label{Tab:JointAccMse}
\centering
\begin{tabular}{l c c c}
\toprule
Method & Classification-Acc & Classification-MSE & Regression-MSE\\
\midrule
\midrule
 Manual Localization  & \textbf{63.6\%} & \textbf{0.706} & \textbf{0.503} \\
 Automatic Localization & 61.9\% & 0.781 & 0.541 \\
 \bottomrule
\end{tabular}
\vspace{-2mm}
\end{table}

\subsubsection{Comparing manual and automatic localization:} We present the classification and regression results obtained using both the manual and the automatic methods for localizing the knee joints in Table \ref{Tab:JointAccMse} and Table \ref{Tab:JointMetrics}. From the results, it is evident that the classification and regression of the knee joint images after automatic localization are comparable with the results after manual localization. 

\begin{table}[ht]
\vspace{-5mm}
\caption{Classification metrics after localizing knee joints.}
\label{Tab:JointMetrics}
\centering
\begin{tabular}{c|ccc|ccc}
\hline
\multirow{2}{*}{Grade} & \multicolumn{3}{c|}{Manual Localization} & \multicolumn{3}{c}{Automatic Localization}\tabularnewline
\cline{2-7} 
 & Precision  & Recall & F1 & Precision  & Recall & F1\tabularnewline
\hline
\hline
0 & 0.66 & 0.87 & 0.75 & 0.64 & 0.88 & 0.74\tabularnewline
1 & 0.39 & 0.06 & 0.10 & 0.33 & 0.02 & 0.04\tabularnewline
2 & 0.52 & 0.60 & 0.56 & 0.50 & 0.57 & 0.53\tabularnewline
3 & 0.75 & 0.72 & 0.73 & 0.73 & 0.73 & 0.73\tabularnewline
4 & 0.78 & 0.78 & 0.78 & 0.75 & 0.66 & 0.70\tabularnewline
\hline 
Mean & 0.60 & 0.64 & 0.59 & 0.57 & 0.62 & 0.56 \\
\hline
\end{tabular}
\vspace{-5mm}
\end{table}

\subsubsection{Comparing joint training with classification only:} From the results  shown in Table \ref{Tab:CompClsf} and \ref{Tab:JointAccMse}, the network trained jointly for classification and regression gives higher multi-class classification accuracy of 63.4\% and lower mean-squared error 0.661 in comparison to the previous network trained only for classification with multi-class classification accuracy 60.3\% and mean-squared error 0.898. Table \ref{Tab:CompJLClsf} shows the precision, recall, $F_{1}$ score, and area under curve (AUC) of the network trained jointly for classification and regression and the network trained only for classification. These results show that the network jointly trained for classification and regression learns a better representation in comparison to the previous network trained only for classification. 

\begin{table}[ht]
\vspace{-5mm}
\caption{Metrics comparing joint training for classification and regression to network trained for classification only.}
\label{Tab:CompJLClsf}
\centering
\begin{tabular}{c | c c c c | c c c c}
\hline
     \multirow{2}{*}{Grade} & \multicolumn{4}{c |}{Joint training for Clsf \& Reg} 
      & \multicolumn{4}{c}{Training for only Clsf} \\
      \cline{2-9}
      & Precision & Recall & $F_{1}$ & AUC & Precision & Recall & $F_{1}$ & AUC \\
\hline
\hline
 0 & 0.68 & 0.80 & 0.74  & 0.87    & 0.63 & 0.82 & 0.71 & 0.83 \\
 1 & 0.32 & 0.15 & 0.20  & 0.71    & 0.25 & 0.04 & 0.06 & 0.66 \\
 2 & 0.53 & 0.63 & 0.58  & 0.82    & 0.47 & 0.57 & 0.51 & 0.78 \\
 3 & 0.78 & 0.74 & 0.76  & 0.96    & 0.76 & 0.71 & 0.73 & 0.94 \\
 4 & 0.81 & 0.75 & 0.78  & 0.99    & 0.78 & 0.77 & 0.77 & 0.99 \\
 \hline
 Mean & 0.61 & 0.63 & 0.61 & {-}   & 0.56 & 0.60 & 0.56 & {-} \\
 \hline
\end{tabular}
\vspace{-5mm}
\end{table}

\subsubsection{Error Analysis:}
\begin{figure}[ht]
  \centering
  \begin{minipage}[b]{0.39\textwidth}
    \includegraphics[width=\textwidth]{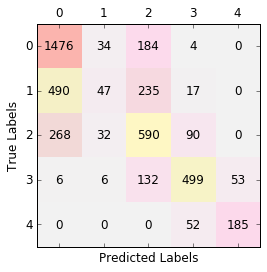}
    \caption{Confusion matrix.}
    \label{fig:ConfMat}
  \end{minipage}
  \hfill
  \begin{minipage}[b]{0.6\textwidth}
    \includegraphics[width=\textwidth]{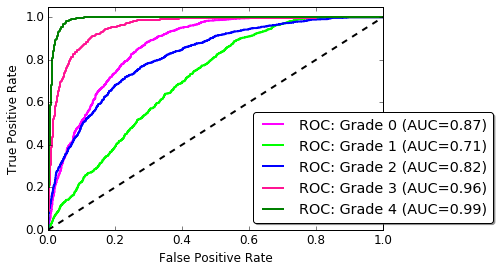}
    \caption{ROC for joint training.}
    \label{fig:ROC}
  \end{minipage}
\end{figure}

From the classification metrics (Table \ref{Tab:CompJLClsf}), the confusion matrix (Figure \ref{fig:ConfMat}) and the receiver operating characteristics (Figure \ref{fig:ROC}), it is evident that classification of successive grades is challenging, and in particular classification metrics for grade 1 have low values in comparison to the other Grades. %\todo[inline]{It's also worth pointing out that most of the errors are made between successive grades. I think including some images of illustrative examples of misclassifications here would be great. I.e. show a typical example of a misclassification from grade 0 to 1 to illustrate how difficult it is to distinguish between these classes. Also it would be good to look at and discuss at least one of the 5 grade 0/3 misclassifications, since these are the most severe. Finding out what went wrong here should give insights.} 

\begin{figure}[ht]
  \centering
  \includegraphics[scale=0.35]{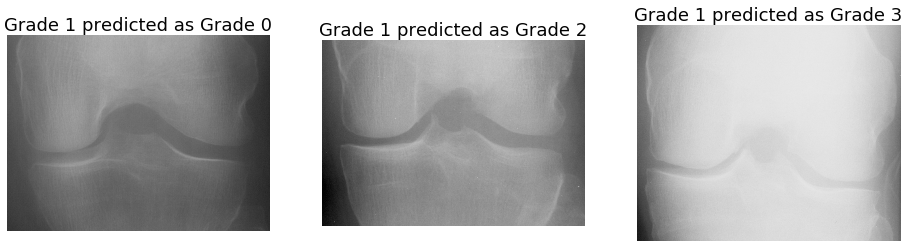}
  \caption{Mis-classifications: grade 1 joints predicted as grade 0, 2, and 3}
  \label{fig:Gr1as023}
%\vspace{-2mm}
\end{figure}

\begin{figure}[ht]
  \centering
  \includegraphics[scale=0.35]{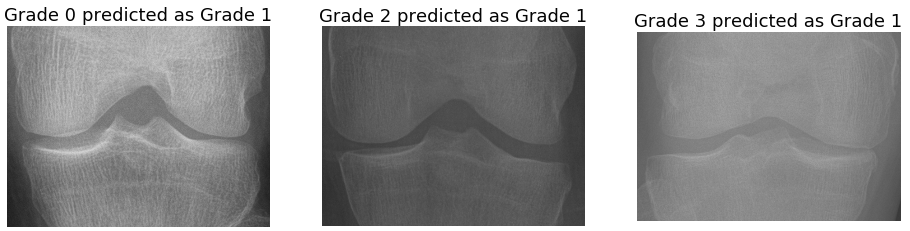}
  \caption{Mis-classifications: other grade knee joints predicted as grade 1}
  \label{fig:Gr023as1}
\vspace{-5mm}
\end{figure}

\begin{figure}[ht]
  \centering
  \includegraphics[scale=0.3]{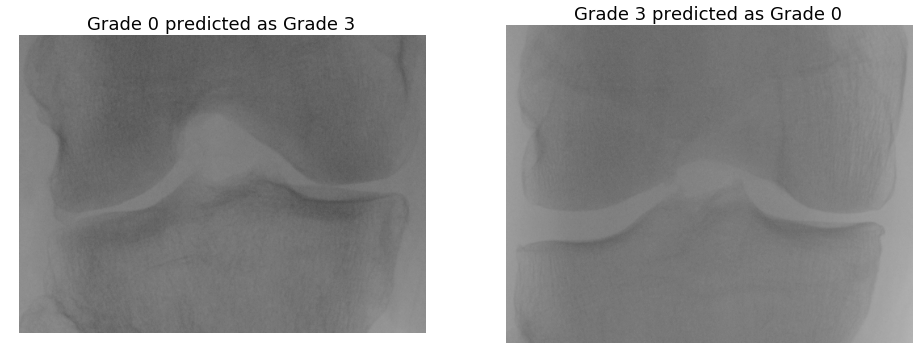}
  \caption{An instance of more severe mis-classification: grade 0 and grade 3}
  \label{fig:Gr03}
\vspace{-8mm}
\end{figure}

Figure \ref{fig:Gr1as023} shows some examples of mis-classifications: grade 1 knee joints predicted as grade 0, 2, and 3. Figure \ref{fig:Gr023as1} shows the mis-classifications of knee joints categorized as grade 0, 2 and 3 predicted as grade 1. These images show minimal variations in terms of joint space width and osteophytes formation, making them challenging to distinguish. Even for the more serious mis-classifications in Figure \ref{fig:Gr03}, e.g. grade 0 predicted as grade 3 and vice versa, do not show very distinguishable variations. 

Even though the KL grades are used for assessing knee OA severity in clinical settings, there has been continued investigation and criticism over the use of KL grades as the individual categories are not equidistant from each other \cite{emrani2008joint,hart2003kellgren}. This could be a reason for the low multi-class classification accuracy in the automatic quantification. Using OARSI readings instead of KL grades could possibly provide better results for automatic quantification as the knee OA features such as joint space narrowing, osteophytes formation, and sclerosis are separately graded. %Moreover, when we visually observe the knee X-ray images belonging to Grade 0 and Grade 1 severity, there are very subtle variations in terms of the joint space width and osteophytes formation. To capture these variations and distinguish coarse grades such as Grade 0 and Grade 1, we require better learning representations.

\begin{comment}
\begin{table}[ht]
\caption{Confusion matrix for classification using jointly trained network.}
\label{Tab:ConfMat}
\centering
\begin{tabular}{c c c c c c c}
\toprule
Grade & 0 & 1 & 2 & 3 & 4 & Total Samples\\
\midrule
 0  & \textbf{1422} & 84 & 191 & 1 & 0 & 1698 \\
 1 & 453 & \textbf{81} & 232 & 23 & 0 & 789 \\
 2 & 231 & 65 & \textbf{606} & 78 & 0 & 980 \\
 3 & 4 & 7 & 125 & \textbf{509} & 51 & 696 \\
 4 & 0 & 0 & 0 & 56 & \textbf{181} & 237 \\
\bottomrule
\end{tabular}
\todo[inline]{Which axis is the true labels and which is the predicted ones?}
\end{table}
\end{comment}
% ---------------------------------- %

\section{Conclusion}
We proposed new methods to automatically localize knee joints using a fully convolutional network  and quantified knee OA severity through a network jointly trained for multi-class classification and regression where both networks were trained from scratch. The FCN based method is highly accurate in comparison to the previous methods. We showed that the classification results obtained with automatically localized knee joints is comparable with the manually segmented knee joints. There is an improvement in the multi-class classification accuracy, precision, recall, and $F_{1}$ score of the jointly trained network for classification and regression in comparison to the previous method. The confusion matrix and other metrics show that classifying Knee OA images conditioned on KL grade 1 is challenging due to the small variations, particularly in the consecutive grades from grade 0 to grade 2. 

Future work will focus on training an end-to-end network to quantify the knee OA severity integrating the FCN for localization and the CNN for classification. It will be interesting to investigate the human-level accuracy involved in assessing the knee OA severity and comparing this to the automatic quantification methods. This could provide insights to further improve fine-grained classification.  

% ---------------------------------- %

\section*{Acknowledgment}

This publication has emanated from research conducted with the financial support of Science Foundation Ireland (SFI) under grant numbers SFI/12/RC/2289 and 15/SIRG/3283.

The OAI is a public-private partnership comprised of five contracts (N01-AR-2-2258; N01-AR-2-2259; N01-AR-2- 2260; N01-AR-2-2261; N01-AR-2-2262) funded by the National Institutes of Health, a branch of the Department of Health and Human Services, and conducted by the OAI Study Investigators. Private funding partners include Merck Research Laboratories; Novartis Pharmaceuticals Corporation, GlaxoSmithKline; and Pfizer, Inc. Private sector funding for the OAI is managed by the Foundation for the National Institutes of Health.

MOST is comprised of four cooperative grants (Felson
-- AG18820; Torner -- AG18832; Lewis -- AG18947; and Nevitt -- AG19069) funded by the National Institutes of Health, a branch of the Department of Health and Human Services, and conducted by MOST study investigators. This manuscript was prepared using MOST data and does not necessarily reflect the opinions or views of MOST investigators.

% ---------------------------------- %

\bibliographystyle{splncs03} 
\bibliography{MLDM_Biblio}

%\nocite{*}

\end{document}